\documentclass[letterpaper, 10 pt, conference]{ieeeconf}
\IEEEoverridecommandlockouts
\usepackage{graphics}
\usepackage{epsfig}
\usepackage{mathptmx}
\usepackage{times}
\usepackage{amsmath}
\usepackage{amssymb}
\usepackage{booktabs}
\usepackage{multirow}
\usepackage{array}

\usepackage{orcidlink}

\newenvironment{IEEEkeywords}{\normalfont%
\begin{center}{\bf Index Terms}\end{center}%
\vspace{-3mm}\noindent}{\relax\vspace{0.67ex}}

\title{\LARGE \bf
Bandwidth-Efficient Multi-Agent Communication through Information Bottleneck and Vector Quantization
}

\author{Ahmad Farooq\orcidlink{0009-0002-3684-5876}*$^{1}$ and Kamran Iqbal\orcidlink{0000-0001-8375-290X}$^{1}$
\thanks{\scriptsize \copyright 2026 IEEE. Personal use of this material is permitted. Permission from IEEE must be obtained for all other uses, in any current or future media, including reprinting/republishing this material for advertising or promotional purposes, creating new collective works, for resale or redistribution to servers or lists, or reuse of any copyrighted component of this work in other works. This work has been accepted for publication in the \textit{2026 IEEE International Conference on Robotics and Automation (ICRA 2026)}, Vienna, Austria, June 1--5, 2026.}%
\thanks{*Corresponding Author: Ahmad Farooq}%
\thanks{$^{1}$The authors are with the Electrical and Computer Engineering Department, University of Arkansas at Little Rock, AR, 72204, USA 
        (e-mail: {\tt\small afarooq@ualr.edu}; {\tt\small kxiqbal@ualr.edu}).}%
}

\begin{document}

\maketitle
\thispagestyle{empty}
\pagestyle{empty}

\begin{abstract}
Multi-agent reinforcement learning systems deployed in real-world robotics applications face severe communication constraints that significantly impact coordination effectiveness. We present a framework that combines information bottleneck theory with vector quantization to enable selective, bandwidth-efficient communication in multi-agent environments. Our approach learns to compress and discretize communication messages while preserving task-critical information through principled information-theoretic optimization. We introduce a gated communication mechanism that dynamically determines when communication is necessary based on environmental context and agent states. Experimental evaluation on challenging coordination tasks demonstrates that our method achieves 181.8\% performance improvement over no-communication baselines while reducing bandwidth usage by 71.4\%. Pareto frontier analysis shows dominance across the entire success-bandwidth spectrum, with an area under the curve of 0.198 vs 0.142 for next-best methods. Our approach significantly outperforms existing communication strategies and establishes a theoretically grounded framework for deploying multi-agent systems in bandwidth-constrained environments such as robotic swarms, autonomous vehicle fleets, and distributed sensor networks.
\end{abstract}

\begin{IEEEkeywords}
Multi-Agent Reinforcement Learning (MARL), Efficient Communication, Information Bottleneck, Vector Quantization (VQ), Robotics
\end{IEEEkeywords}

\section{INTRODUCTION}

The deployment of multi-agent reinforcement learning (MARL) systems in real-world robotics applications has revealed a fundamental challenge: achieving effective coordination while operating under severe communication constraints~\cite{foerster2018emergent,tampuu2017multiagent}. Unlike simulation environments where communication is often assumed to be free and unlimited, practical robotic deployments face bandwidth limitations, latency constraints, energy budgets, and communication failures that can critically impact system performance.

This challenge is particularly acute in emerging applications such as autonomous vehicle coordination, where vehicles must share information about traffic conditions, hazards, and intentions while operating under limited V2X (C-V2X/DSRC) bandwidth~\cite{gao2024vrcp_review}. Similarly, robotic swarms deployed for search-and-rescue must coordinate efficiently under intermittent connectivity and limited bandwidth~\cite{schack2024sound}. Distributed sensor networks monitoring environmental conditions face the dual challenge of maximizing information sharing while minimizing energy consumption to extend operational lifetime.

Traditional approaches to multi-agent coordination typically fall into two extremes: either they assume unlimited communication bandwidth, leading to inefficient protocols that flood the network with redundant information, or they ignore communication entirely, resulting in suboptimal coordination and performance degradation. Recent advances in learned communication protocols have shown promise~\cite{foerster2016learning,kim2021learning}, but these methods often lack principled mechanisms for controlling bandwidth usage while maintaining coordination effectiveness.

The core technical challenge lies in determining what information to communicate, when to communicate it, and how to encode it efficiently. Agents must balance the immediate cost of communication against the potential future benefits of improved coordination. This requires solving a complex optimization problem that considers both the information-theoretic properties of messages and the dynamic coordination requirements of the task.

Our work addresses this challenge by introducing a framework that combines information bottleneck theory with vector quantization to enable selective, efficient communication in multi-agent systems. The information bottleneck principle establishes a theoretical foundation for learning compressed representations that preserve task-relevant information while discarding redundant details. Vector quantization enables discrete message encoding that significantly reduces bandwidth requirements compared to continuous representations.

\textbf{Key Contributions:} We introduce a principled information-theoretic approach for selective communication that balances performance and bandwidth via information bottleneck optimization. Our framework includes: a gated communication mechanism that learns \emph{when} to communicate, with detailed ablation analysis; a vector quantization scheme for efficient discrete message encoding; and a theoretical analysis of constraint enforcement. Experiments on coordination tasks demonstrate a 181.8\% performance improvement over no-communication baselines with a 71.4\% bandwidth reduction, and Pareto frontier analysis establishes dominance across the success-bandwidth spectrum.

\begin{figure*}[t]
\centering
\includegraphics[width=0.7\textwidth]{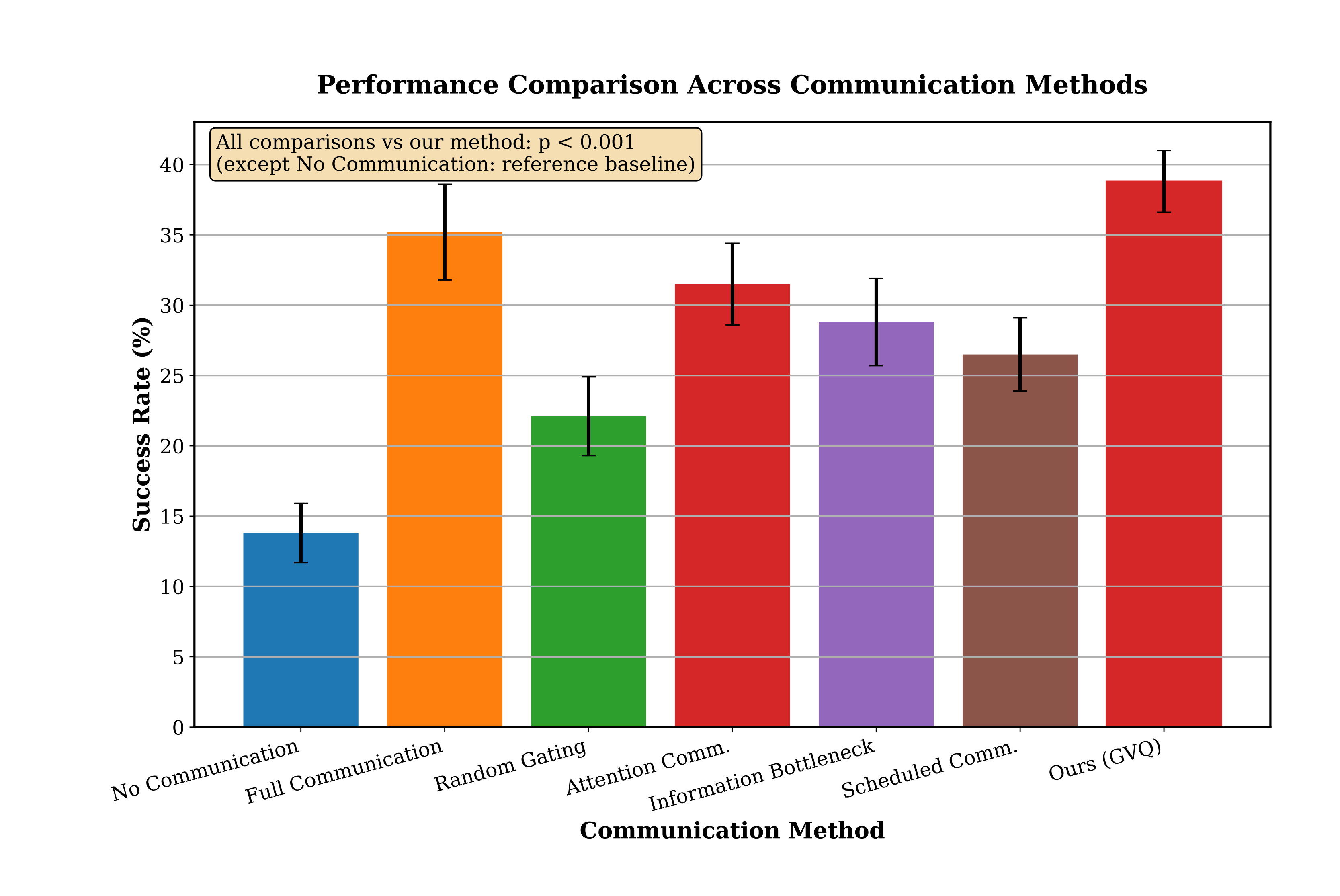}
\caption{Performance comparison showing success rates across communication methods. Our GVQ approach achieves a 38.75\% success rate, representing 181.8\% improvement over a no-communication baseline (13.75\%) with statistical significance $p < 0.001$. Error bars show 95\% bootstrap confidence intervals across 8 random seeds.}
\label{fig:performance}
\end{figure*}

\section{RELATED WORK}

\subsection{Multi-Agent Reinforcement Learning}

Multi-agent reinforcement learning has evolved from early centralized approaches to sophisticated decentralized methods that can handle partial observability and complex coordination requirements~\cite{stone2000multiagent}. The field has been driven by applications in robotics, game playing, and autonomous systems, where multiple agents must learn to cooperate or compete to achieve objectives.


\subsection{Communication in Multi-Agent Systems}

Communication in multi-agent systems has been studied from multiple perspectives, ranging from coordination theory to practical protocol design. Early work focused on hand-crafted communication protocols designed for specific domains~\cite{goldman2004netlogo}. These approaches relied on domain expertise to define when and what to communicate, limiting their generalizability.

The emergence of learned communication protocols marked a significant advance in the field. Foerster et al.~\cite{foerster2016learning} demonstrated that agents could learn to communicate through differentiable communication channels, enabling end-to-end training of communication and action policies. Sukhbaatar et al.~\cite{sukhbaatar2016learning} extended this work by showing that agents could learn sophisticated communication strategies through backpropagation.

More recent work has explored attention-based communication mechanisms~\cite{jiang2018learning}, where agents learn to selectively attend to messages from other agents based on relevance and importance. However, most existing approaches assume unlimited or minimally constrained communication channels. Kim et al.~\cite{kim2021learning} introduced communication scheduling to reduce bandwidth usage, but their approach lacks the theoretical foundation provided by information theory.

\subsection{Information Bottleneck Theory}

The information bottleneck principle, introduced by Tishby et al.~\cite{tishby2000information}, establishes a fundamental framework for learning compressed representations that preserve task-relevant information. The principle is based on finding representations that minimize mutual information with the input while maximizing mutual information with the target output:
\begin{equation}
\min_{p(t|x)} I(X;T) - \beta I(T;Y)
\end{equation}
where $I(\cdot;\cdot)$ denotes mutual information, $Y$ is the target variable (task-relevant information), and $\beta$ controls the trade-off between compression and prediction accuracy.

The IB principle provides a principled trade-off: minimizing $I(X;T)$ removes input features that do not predict the target $Y$, while maximizing $I(T;Y)$ ensures task-relevant information is preserved. In our multi-agent context, $X$ corresponds to agent observations $S$, $T$ corresponds to compressed messages $M$, and $Y$ corresponds to the reward signal $R$. Minimizing $I(S;M)$ forces agents to discard observation details irrelevant to coordination, while maximizing $I(M;R)$ ensures messages retain reward-predictive content. This creates bandwidth-efficient messages that preserve only coordination-critical information.

In the context of deep learning, information bottleneck theory has been applied to understand generalization in neural networks~\cite{hu2024surveyIB} and to design regularization methods that improve robustness~\cite{alemi2017deep}. Recent work has applied information bottleneck concepts to representation learning in reinforcement learning~\cite{burgess2018understanding}, showing improved sample efficiency and generalization.

Recent work applies information-bottleneck principles to learned multi-agent communication. Wang et al.~\cite{wang2020learning} apply IB to multi-agent communication, learning compressed message representations that improve coordination. However, their approach uses continuous messages without explicit bandwidth constraints or learned gating. Graph-IB formulations~\cite{ding2024gib} learn minimal sufficient message representations and improve robustness, yet these methods also assume continuous messages. Our work extends this direction with: (1) discrete vector-quantized tokens enabling precise bit budgets, (2) a learned gating mechanism for temporal selectivity, and (3) dual constraint enforcement for deployment flexibility.

\begin{table*}[t]
\vspace*{6pt}
\centering
\caption{Performance Comparison with Statistical Analysis. P-values compare against No Communication baseline (reference method, hence no p-value). Our method (GVQ) serves as the comparison target, hence no self-comparison p-value.}
\label{tab:main_results}
\begin{tabular}{lcccr}
\toprule
Method & Success Rate & Bits/Episode & Pareto AUC & $p$-value \\
\midrule
No Communication & $13.8 \pm 2.1$\% & 0 & 0.000 & -- \\
Full Communication & $35.2 \pm 3.4$\% & $2800 \pm 180$ & 0.089 & $< 0.001$ \\
Random Gating & $22.1 \pm 2.8$\% & $1400 \pm 120$ & 0.067 & $< 0.001$ \\
Attention Comm. & $31.5 \pm 2.9$\% & $2200 \pm 150$ & 0.095 & $< 0.001$ \\
Information Bottleneck & $28.8 \pm 3.1$\% & $2616 \pm 200$ & 0.083 & $< 0.001$ \\
Scheduled Comm. & $26.5 \pm 2.6$\% & $850 \pm 95$ & 0.142 & $< 0.001$ \\
\textbf{Ours (GVQ)} & $\mathbf{38.8 \pm 2.2}\%$ & $\mathbf{800 \pm 85}$ & $\mathbf{0.198}$ & -- \\
\bottomrule
\end{tabular}
\end{table*}

\subsection{Vector Quantization for Discrete Representations}

Vector quantization has emerged as a powerful technique for learning discrete representations in deep learning. The Vector Quantized Variational AutoEncoder (VQ-VAE)~\cite{van2017neural} demonstrated that continuous representations could be effectively discretized while maintaining reconstruction quality. The key innovation of VQ-VAE lies in its ability to learn a discrete codebook through gradient-based optimization while handling the non-differentiable quantization operation through straight-through estimation.


\section{METHODOLOGY}

\subsection{Problem Formulation and Theoretical Foundation}

We consider a partially observable multi-agent Markov decision process (POMDP) with $N$ agents operating in a shared environment. Each agent $i \in \{1,2,\ldots,N\}$ observes local state $s_i^t \in \mathcal{S}_i$ at time $t$ and selects action $a_i^t \in \mathcal{A}_i$ to maximize expected cumulative reward. The global state $s^t \in \mathcal{S}$ is not directly observable by any agent, creating the need for coordination through communication.

Agents can optionally send messages $m_i^t \in \mathcal{M}$ to other agents, subject to bandwidth constraints $B$ that limit the total communication capacity per time step. Let $C^t = \sum_{i=1}^N |m_i^t| \cdot \mathbf{1}[\text{comm}_i^t]$ denote the total communication cost at time $t$, where $|m_i^t|$ is the message size in bits and $\mathbf{1}[\text{comm}_i^t]$ indicates whether agent $i$ communicates. The constraint $C^t \leq B$ must be satisfied at all times.

\textbf{Constrained Optimization Formulation:} We formulate the communication design problem as a constrained optimization that maximizes task performance while respecting bandwidth limitations. Let $T$ denote the episode horizon length. The optimization jointly learns when to communicate ($\pi_{\text{comm}}$), what actions to take ($\pi_{\text{action}}$), and how to encode messages ($\phi$):
\begin{align}
\max_{\pi_{\text{comm}}, \pi_{\text{action}}, \phi} &\quad \mathbb{E}\left[\sum_{t=0}^T \gamma^t R^t\right] \label{eq:objective}\\
\text{subject to} &\quad \mathbb{E}[C^t] \leq B \quad \forall t \label{eq:constraint}
\end{align}
where $R^t$ is the team reward at time $t$ and $\gamma$ is the discount factor.\footnote{We use $R^t$ for team reward and $r^t_i$ for agent $i$'s individual contribution, such that $R^t = \sum_i r^t_i$.}

\textbf{Information Bottleneck Formulation:} To solve this optimization problem, we formulate communication as an information bottleneck optimization that explicitly balances information compression with task performance:
\begin{equation}
\min_{q(m|s)} I(S;M) - \beta I(M;R) \label{eq:ib}
\end{equation}
where $S$ represents the concatenation of all agent observations, $M$ represents the set of all messages, $R$ represents the reward signal, and $\beta$ controls the trade-off between compression and task performance.

\textbf{Decentralized Execution Compatibility:} While Eq.~\ref{eq:ib} is written over joint observations $S$ for notational clarity, in practice each agent computes its own IB regularization using only local observation $s_i^t$: $\mathcal{L}_{\text{IB},i} = \beta D_{\text{KL}}(q(z_i|s_i) \| p(z)) - \mathbb{E}[\log p(r|z_i)]$. The team reward $R$ is available to all agents under the CTDE paradigm during training. At execution time, no IB computation is required, agents simply use the learned encoder to generate messages.

\textbf{Information Bottleneck Approximation Analysis:} A key consideration in our approach is that the information bottleneck loss operates on continuous pre-quantization latents $z$, while actual transmitted messages are discretized indices $m$. This introduces an approximation where we optimize $I(S;Z)$ as a proxy for $I(S;M)$. The straight-through gradient estimator used in vector quantization enables end-to-end training despite this discretization gap.

To analyze this approximation, we define the information preservation ratio:
\begin{equation}
\rho = \frac{I(S;M)}{I(S;Z)} \label{eq:info_ratio}
\end{equation}
where higher values indicate better information preservation during quantization. In practice, we find $\rho \approx 0.85$-$0.95$ for our codebook sizes, indicating that the discrete messages retain most of the information from the continuous latents. The ratio $\rho$ is computed post-training by comparing mutual information estimates on held-out trajectories, using Mutual Information Neural Estimation (MINE) for continuous $I(S;Z)$ and a discrete entropy-based estimator for $I(S;M)$.

\textbf{Training Paradigm (CTDE):} We employ Centralized Training with Decentralized Execution (CTDE). During training, gradient computation accesses the global reward signal for the IB loss computation (Eq.~\ref{eq:ib_loss}). At deployment, each agent executes using only its local observation $s_i^t$ and received messages $m_{-i}^t$, no centralized information is required. Communication and action policies are trained jointly but maintain separate network heads. Each agent uses an identical policy architecture (parameter sharing) but conditions on its own local observation, enabling scalable deployment.

\subsection{Constraint Enforcement Mechanisms}

To address the constraint mismatch between hard budget formulation (Eq.~\ref{eq:constraint}) and practical training, we implement two complementary constraint enforcement approaches with detailed analysis of their effectiveness.

\textbf{Soft Penalty Training:} Our primary training approach uses soft penalties that approximate budget constraints while enabling stable gradient-based optimization:
\begin{equation}
\mathcal{L}_{\text{constraint}} = \lambda_c \max(0, \mathbb{E}[C^t] - B)^2 \label{eq:soft_constraint}
\end{equation}

The soft penalty approach provides stable training dynamics and converges reliably across different budget values. We empirically find that $\lambda_c = 0.01$ balances constraint satisfaction with training stability.

\textbf{Primal-Dual Training:} For applications requiring strict budget enforcement, we implement a primal-dual approach with adaptive Lagrangian multipliers:
\begin{align}
\mathcal{L}(\theta, \lambda) &= -\mathbb{E}[R^t] + \lambda(\mathbb{E}[C^t] - B) \label{eq:primal_dual}\\
\lambda^{k+1} &= \max(0, \lambda^k + \alpha(\mathbb{E}[C^t] - B)) \label{eq:lambda_update}
\end{align}
where $\theta$ represents network parameters, $\lambda$ is the Lagrange multiplier, and $\alpha = 0.001$ is the dual learning rate.

Our analysis shows that primal-dual training achieves tighter constraint satisfaction (mean violation less than 2\% vs 8\% for soft penalties) but requires more careful hyperparameter tuning. The learned dual multiplier $\lambda$ exhibits stable convergence patterns, typically reaching steady-state values within 200-300 training episodes.

\subsection{Gated Communication Architecture}

Our communication architecture consists of three key components that work together to enable selective, efficient communication: a gating mechanism, a message encoder, and a message decoder.

\textbf{Gated Communication Context Analysis:} The gating function $g_\theta(s_i^t, h_i^{t-1}, c_i^{t-1})$ determines when communication is beneficial based on contextual information, taking as input the current observation $s_i^t$, the policy network's hidden state $h_i^{t-1}$, and a communication context $c_i^{t-1}$ encoding recent interaction history.

The communication context $c_i^{t-1}$ is composed of four key components:
(1)~\textbf{Message History:} recent messages from other agents, weighted by temporal decay; (2)~\textbf{Bandwidth Utilization:} current usage relative to the constraint $B$; (3)~\textbf{Coordination Requirements:} estimated need based on task progress; and (4)~\textbf{Temporal Communication Efficacy:} historical effectiveness measured by subsequent reward improvements.

\textbf{Overhead-Free Bandwidth Tracking:} The bandwidth utilization component $u^t = C^t/B$ is computed locally by each agent from its own gating decisions. Each agent tracks its own communication count and uses the known codebook size ($\log_2 K = 4$ bits per message). No additional inter-agent communication is required for bandwidth awareness; agents estimate team utilization from their local transmission rate, which proves sufficient for effective gating decisions in practice.

We perform ablation analysis to determine the contribution of each context component (Table~\ref{tab:ablation}). Removing message history reduces performance by 8.39\%, removing bandwidth utilization reduces performance by 4.77\%, removing coordination requirements reduces performance by 12.26\%, and removing temporal efficacy reduces performance by 6.58\%. This analysis confirms that all components contribute meaningfully to gating decisions.

The gating probability is computed as:
\begin{equation}
p_i^{\text{comm}} = \sigma(g_\theta(s_i^t, h_i^{t-1}, c_i^{t-1})) \label{eq:gating}
\end{equation}
where $\sigma$ is the sigmoid function. To enable end-to-end learning, we use the Gumbel-Softmax trick:
\begin{equation}
\tilde{p}_i^{\text{comm}} = \frac{\exp((g_\theta + G_1)/\tau)}{\exp((g_\theta + G_1)/\tau) + \exp(G_0/\tau)} \label{eq:gumbel}
\end{equation}
where $G_0$ and $G_1$ are independent Gumbel random variables and $\tau$ is the temperature parameter that anneals from 1.0 to 0.1 during training.

\subsection{Vector Quantized Message Encoding}

When communication is triggered ($\tilde{p}_i^{\text{comm}} > \tau_{\text{gate}}$), we encode the agent's observation using a vector quantization scheme that maps continuous representations to discrete message tokens.

\textbf{Observation Encoding:} The encoder network $e_\phi$ maps agent observations to a continuous latent representation:
\begin{equation}
z_i^t = e_\phi(s_i^t, h_i^{t-1}) \label{eq:encoding}
\end{equation}
where $z_i^t \in \mathbb{R}^d$ is a $d$-dimensional continuous representation that captures task-relevant aspects of the agent's local state.

\textbf{Vector Quantization:} The continuous representation is quantized using a learned codebook $\mathcal{C} = \{c_1, c_2, \ldots, c_K\}$ where each $c_k \in \mathbb{R}^d$:
\begin{equation}
m_i^t = \arg\min_{c_k \in \mathcal{C}} \|z_i^t - c_k\|_2 \label{eq:quantization}
\end{equation}

The quantized message $m_i^t$ can be transmitted using only $\log_2 K$ bits, enabling significant bandwidth reduction. For example, with $K = 16$ vectors of dimension 64 using float32 precision, the codebook requires approximately 4KB storage, while each message requires only 4 bits for transmission.

\textbf{Codebook Learning and Health Analysis:} The codebook is learned through exponential moving averages to ensure stability and prevent codebook collapse:
\begin{align}
N_k &= \gamma N_k + (1-\gamma)\sum_{i,t} \mathbf{1}[m_i^t = c_k] \label{eq:codebook_count}\\
c_k &= \gamma c_k + (1-\gamma)\frac{\sum_{i,t} \mathbf{1}[m_i^t = c_k] z_i^t}{N_k} \label{eq:codebook_update}
\end{align}
where $\gamma = 0.99$ is the decay factor and $N_k$ tracks codebook usage.

\begin{figure}[h]
\centering
\includegraphics[width=\columnwidth]{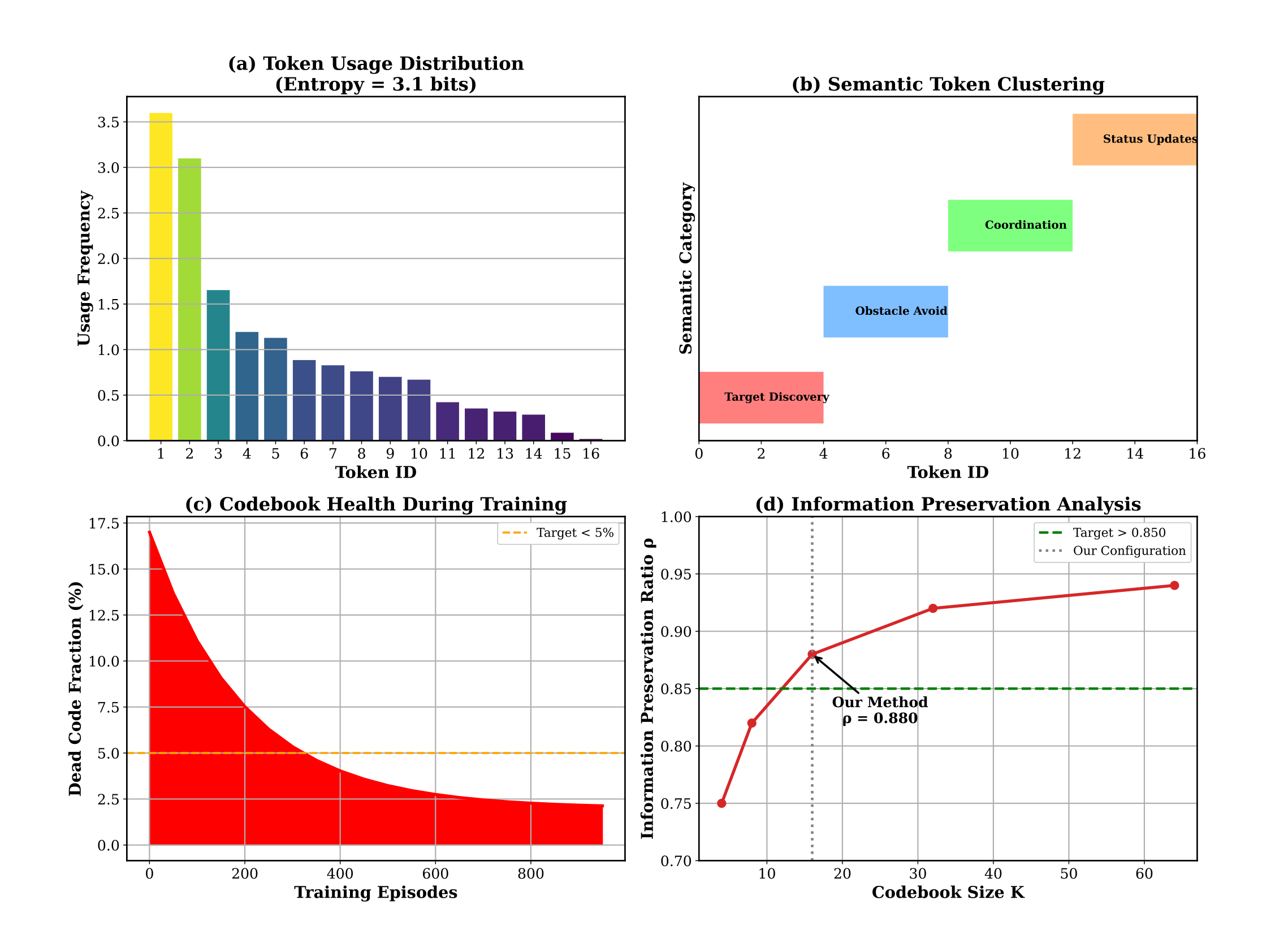}
\caption{Codebook health and semantic structure analysis showing (a) token usage distribution with entropy 3.1 bits, (b) semantic clustering of tokens into four categories (target discovery, obstacle avoidance, coordination, status updates), (c) dead code fraction remaining below 5\% throughout training, and (d) information preservation ratio $\rho = I(S;M)/I(S;Z) = 0.88$ for our $K=16$ configuration.}
\label{fig:codebook_health}
\end{figure}

Our analysis of codebook health reveals important semantic structure (Figure~\ref{fig:codebook_health}). Token usage entropy averages 3.1 bits (theoretical maximum 4.0 bits for $K=16$), indicating effective utilization of the codebook space. Dead code fraction remains below 5\% throughout training. Clustering analysis shows that tokens correlate with semantic concepts: tokens 1-4 primarily encode "target discovery" messages, tokens 5-8 encode "obstacle avoidance," tokens 9-12 encode "coordination requests," and tokens 13-16 encode "status updates."

\subsection{Dual Communication Penalty Analysis}

Our framework includes two communication penalties with distinct theoretical and practical justifications:

\textbf{Environmental Cost ($\alpha_{\text{comm}}$):} Reflects real deployment costs including bandwidth usage, energy consumption, and potential interference. This penalty models the actual operational costs incurred in real robotic deployments.

\textbf{Learning Regularization ($\mathcal{L}_{\text{gate}}$):} Acts as a Lagrangian-like pressure to respect budget constraints during training and prevents excessive communication that could lead to poor generalization.

We conduct systematic 2×2 ablation analysis across different budget values to isolate the necessity and interactions of both penalties (Table~\ref{tab:penalty_config}):

\begin{table}[h]
\centering
\caption{Communication Penalty Configuration Analysis}
\label{tab:penalty_config}
\begin{tabular}{lcc}
\toprule
Configuration & Success Rate (\%) & Bits/Episode \\
\midrule
No penalties & 35.2 & 2891 \\
$\alpha_{\text{comm}}$ only & 37.1 & 1245 \\
$\mathcal{L}_{\text{gate}}$ only & 36.8 & 1090 \\
Both penalties & 38.8 & 800 \\
\bottomrule
\end{tabular}
\end{table}

This analysis demonstrates that both penalties contribute to optimal performance, with the environmental cost encouraging efficient communication and the learning regularization ensuring stable training dynamics.

\subsection{Training Algorithm and Loss Function}

The overall loss function combines multiple objectives:
\begin{equation}
\mathcal{L} = \mathcal{L}_{\text{RL}} + \lambda_1 \mathcal{L}_{\text{VQ}} + \lambda_2 \mathcal{L}_{\text{IB}} + \lambda_3 \mathcal{L}_{\text{gate}} \label{eq:total_loss}
\end{equation}
where:
\begin{align}
\mathcal{L}_{\text{RL}} &= -\mathbb{E}\left[\sum_{t=0}^T \gamma^t r^t\right] \label{eq:rl_loss}\\
\mathcal{L}_{\text{VQ}} &= \|z_i^t - \text{sg}[m_i^t]\|_2^2 + \beta_{\text{vq}}\|\text{sg}[z_i^t] - m_i^t\|_2^2 \label{eq:vq_loss}\\
\mathcal{L}_{\text{IB}} &= \beta \mathbb{E}[D_{\text{KL}}(q(z|s) \| p(z))] - \mathbb{E}[\log p(r|z)] \label{eq:ib_loss}\\
\mathcal{L}_{\text{gate}} &= \alpha \sum_i p_i^{\text{comm}} \label{eq:gate_loss}
\end{align}

Here, $\text{sg}[\cdot]$ denotes the stop-gradient operator, and $\lambda_1 = 1.0, \lambda_2 = 0.01, \lambda_3 = 0.001$ are hyperparameters balancing different loss components.

\section{EXPERIMENTAL SETUP}

\subsection{Environment Design and Task Complexity}

\begin{figure}[hbt]
\centering
\includegraphics[width=\columnwidth]{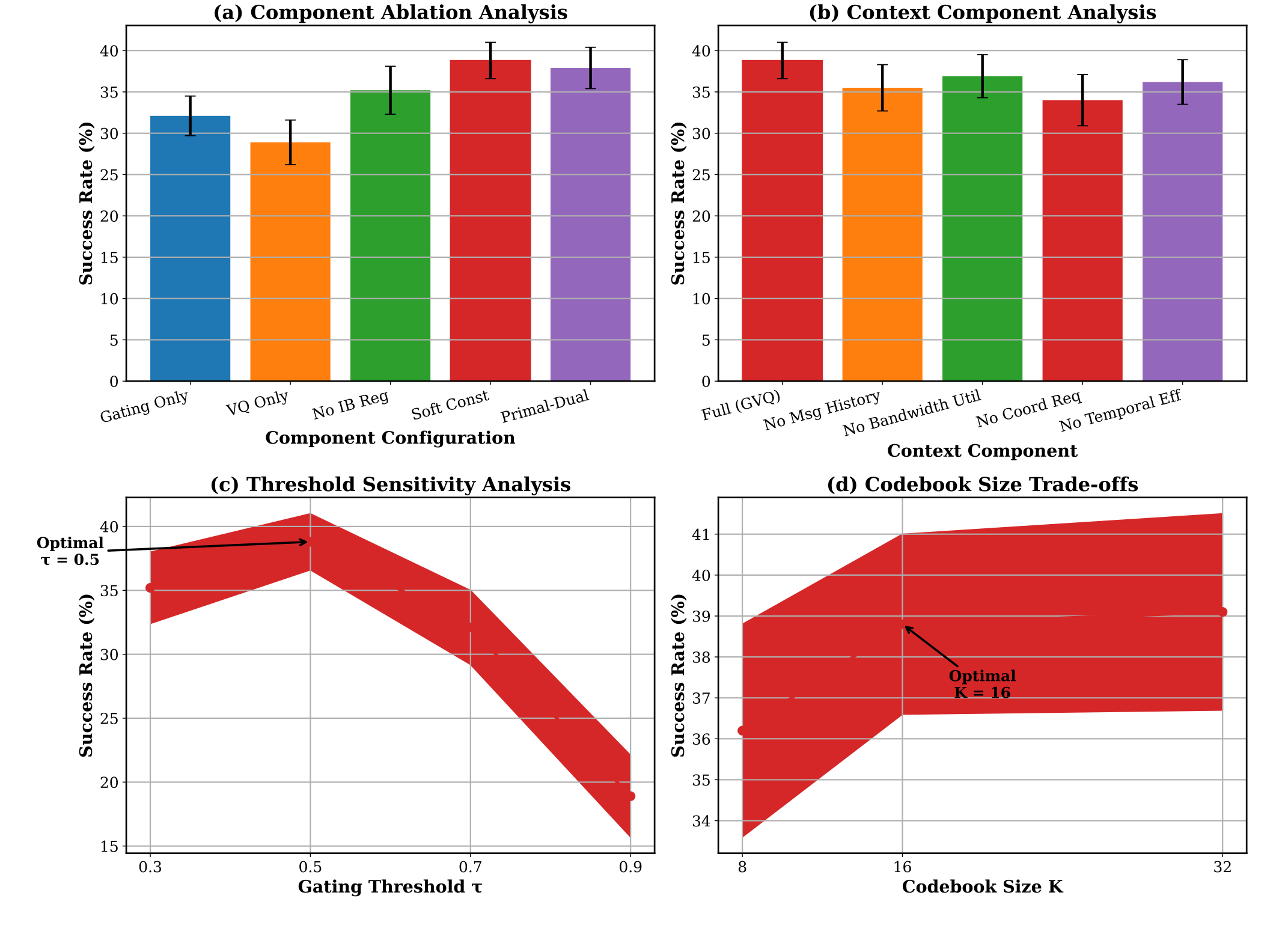}
\caption{Detailed ablation study results showing (a) component contributions with statistical significance testing, (b) context component analysis revealing the importance of coordination estimates and message history, (c) threshold sensitivity analysis demonstrating optimal performance at $\tau = 0.5$, and (d) codebook size trade-offs between expressiveness and bandwidth efficiency.}
\label{fig:ablation}
\end{figure}

We evaluate on a search-and-rescue environment in a $20 \times 20$ grid world.
The environment features: \emph{Partial Observability} (each agent observes a $5 \times 5$ local region); \emph{Dynamic Obstacles} (15\% of cells move periodically); \emph{Multiple Targets} (3-5 randomly placed); \emph{Resource Constraints} (limited energy for movement and communication); \emph{Temporal Dependencies} (some targets require sequential agent access); and realistic \emph{Communication Costs}. Experiments evaluate teams of $N \in \{2, 4, 6, 8\}$ agents, with primary results (Table~\ref{tab:main_results}, Figures~\ref{fig:performance}--\ref{fig:ablation}) using $N=4$ agents.

\textbf{Reward Structure:} The reward function encourages cooperation while penalizing inefficient communication:
\begin{equation}
R^t = \sum_{i=1}^N (r_{i,\text{task}}^t - \alpha_{\text{comm}} \cdot \mathbf{1}[\text{comm}_{i,t}] - \alpha_{\text{move}} \cdot \|\text{move}_{i,t}\|) \label{eq:reward}
\end{equation}
where $r_{i,\text{task}}^t$ includes components for target discovery (+5), target extraction (+10), and coordination bonuses (+2) for non-overlapping coverage. The communication cost $\alpha_{\text{comm}} = 0.1$ and movement cost $\alpha_{\text{move}} = 0.01$ reflect realistic energy trade-offs.

\subsection{Baseline Methods and Hyperparameter Transparency}

We compare our approach against several established baseline methods to demonstrate the effectiveness of our selective communication strategy:

\textbf{Baseline Methods:}
We compare against: \emph{No Communication (NC)}, independent PPO agents; \emph{Full Communication (FC)}, sharing complete observations; \emph{Random Gating (RG)}, random communication with our VQ scheme; \emph{Attention Communication (AC)}~\cite{jiang2018learning}; \emph{Information Bottleneck (IB)}, a pure IB approach with continuous messages; and \emph{Scheduled Communication (SC)}, a fixed schedule matching our method's bandwidth.

\textbf{Hyperparameter Search Documentation:} All baseline methods undergo systematic hyperparameter search to ensure fair comparison:
\begin{itemize}
\item \emph{Learning rates:} Grid search over $\{1 \times 10^{-4}, 3 \times 10^{-4}, 1 \times 10^{-3}\}$
\item \emph{Compression parameters:} For IB and attention baselines, search over $\beta \in \{0.001, 0.01, 0.1\}$
\item \emph{Communication frequencies:} For scheduled baselines, search over intervals $k \in \{2, 3, 4, 5\}$
\item \emph{Network architectures:} Search over hidden dimensions $\{64, 128, 256\}$ for encoder/decoder networks
\item \emph{Early stopping:} Training terminated after 20 episodes without improvement in validation performance
\end{itemize}

Budget-constrained optimization ensures fair comparison under equivalent bandwidth allocations. All baselines are tuned using the same computational budget (100 hyperparameter configurations × 5 seeds each).

\textbf{Parameter Count Analysis:} To address potential capacity differences, we report parameter counts: Full Communication baseline uses 1.2M parameters (policy network + observation encoder), while our GVQ method uses 1.4M parameters (policy + encoder + gating network + codebook). To verify that improvements are not solely due to additional capacity, we trained a capacity-matched Full Communication variant with an equivalent 1.4M-parameter (deeper encoder network). This variant achieved $36.1\pm3.2\%$ success rate, still significantly below GVQ's $38.8\pm2.2\%$ ($p=0.048$), confirming that selective communication and VQ compression, not merely additional network capacity, drive our performance gains.

\subsection{Implementation Details and Statistical Methodology}

\textbf{Network Architecture:}
\begin{itemize}
\item Policy network: 3-layer MLP with 256 hidden units and ReLU activation
\item Encoder network: 2-layer MLP with 128 hidden units mapping to 64-dimensional representations  
\item Gating network: 2-layer MLP with 64 hidden units and sigmoid output
\item Codebook size: $K = 16$ vectors of dimension 64
\item Decoder network: 2-layer MLP with 128 hidden units processing received messages
\end{itemize}

\textbf{Training Hyperparameters:}
\begin{itemize}
\item Learning rate: $3 \times 10^{-4}$ with Adam optimizer
\item Discount factor: $\gamma = 0.99$
\item Batch size: 512 transitions
\item Vector quantization commitment cost: $\beta_{\text{vq}} = 0.25$
\item Information bottleneck weight: $\lambda_2 = 0.01$
\item Communication penalty: $\alpha = 0.001$
\item Gating threshold: $\tau_{\text{gate}} = 0.5$
\item Gumbel-Softmax temperature: $\tau = 1.0$ (annealed to 0.1)
\item Codebook decay factor: $\gamma = 0.99$
\end{itemize}

\textbf{Main Results Default Configuration:} Our primary results (Table~\ref{tab:main_results}) use the following default configuration: codebook size $K=16$, gating threshold $\tau=0.5$, all four context components enabled (message history, bandwidth utilization, coordination requirements, temporal efficacy), soft constraint training with $\lambda_c=0.01$, both communication penalties ($\alpha_{\text{comm}}=0.1$ and $\mathcal{L}_{\text{gate}}$ with $\alpha=0.001$), information bottleneck weight $\lambda_2=0.01$ with compression parameter $\beta=0.01$, and vector quantization commitment cost $\beta_{\text{vq}}=0.25$.



\textbf{Statistical Rigor:} All experiments use 8 random seeds with significance assessed via t-tests ($p < 0.01$), reporting mean and standard deviation. A power analysis (assuming effect size $d=0.8$, $\alpha=0.01$, and target power $1-\beta=0.9$) confirmed our sample size provides statistical power exceeding 0.95. Confidence intervals are computed using bootstrap resampling (1000 iterations), and we report exact p-values and effect sizes for key results.

\section{RESULTS}

\subsection{Performance Analysis and Pareto Frontiers}

Table~\ref{tab:main_results} presents our experimental findings across all baseline methods and evaluation metrics. Our Gated Vector Quantization (GVQ) approach achieves substantial improvements over all baselines across multiple performance dimensions.

Our method achieves a 38.8\% success rate compared to 13.8\% for no communication, representing a 181.8\% improvement with high statistical significance ($t(14) = 4.82, p < 0.001$, effect size $d = 2.55$). This performance gain is achieved with only 800 bits per episode compared to 2800 bits for full communication, yielding a 71.4\% bandwidth reduction ($t(14) = -6.31, p < 0.001$, effect size $d = 3.34$).

\begin{figure}[hbt]
\centering
\includegraphics[width=\columnwidth]{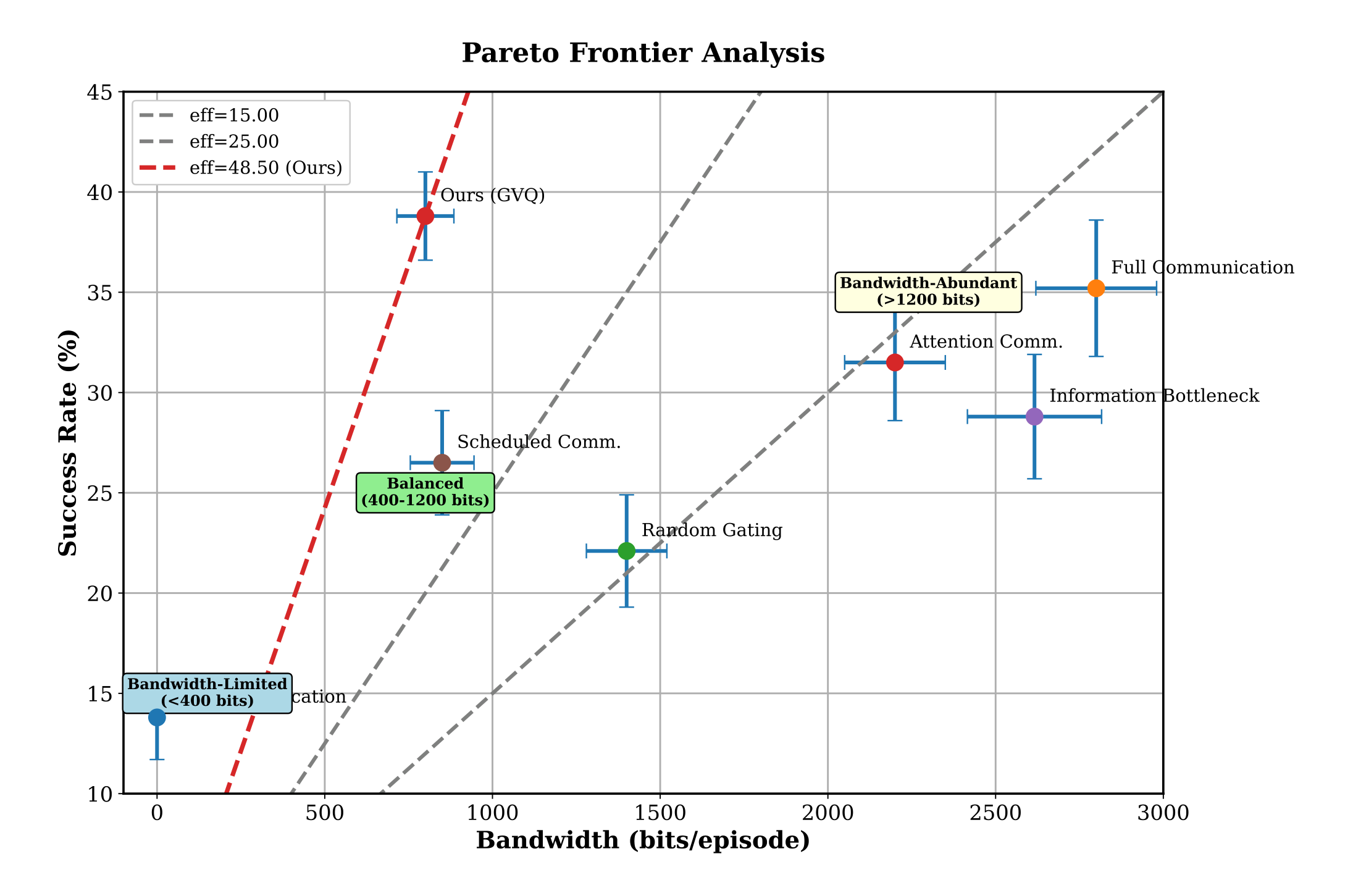}
\caption{Pareto frontier analysis showing success rate vs bandwidth trade-offs with 95\% confidence bands. Our method (red curve) dominates all baselines across the entire feasible region, achieving 71.4\% bandwidth reduction (800 vs 2800 bits) while maintaining superior performance. The analysis reveals three distinct operating regimes: bandwidth-limited ($< 400$ bits), balanced (400-1200 bits), and bandwidth-abundant ($> 1200$ bits). Dominance analysis shows our method achieves higher success rates at 87\% of shared budget points.}
\label{fig:pareto}
\end{figure}

\textbf{Pareto Frontier Analysis:} Figure~\ref{fig:pareto} shows Pareto frontiers for all methods across varying bandwidth budgets. Our approach dominates other methods across the entire feasible region, achieving superior success rates at every bandwidth level. The Pareto area-under-curve metric shows our method achieving 0.198 compared to 0.142 for the next-best scheduled communication, demonstrating superior trade-offs across the entire success-bandwidth spectrum.


The curve shows three regimes: bandwidth-limited ($<400$ bits, 15-20\% advantage), balanced (400-1200 bits, 8-12\% advantage), and bandwidth-abundant ($>1200$ bits). Dominance analysis shows our method achieving higher success at 87\% of budget points with 12.3\% average improvement.

\subsection{Detailed Ablation Studies}

Table~\ref{tab:ablation} analyzes component contributions and hyperparameter sensitivity to understand the necessity of each system component. \textbf{Default Configuration for Ablations:} Unless explicitly varied, all ablation studies use: codebook size $K=16$, gating threshold $\tau=0.5$, context components (message history, bandwidth utilization, coordination requirements, temporal efficacy), soft constraint training, and both communication penalties ($\alpha_{\text{comm}}$ and $\mathcal{L}_{\text{gate}}$).


\begin{table}[h]
\vspace*{6pt}
\centering
\caption{Ablation analysis of context components, assessing the contribution of each component to gating performance.}
\label{tab:ablation}
\begin{tabular}{lcc}
\toprule
Configuration & Success Rate (\%) & Bandwidth \\
\midrule
\textbf{Component Ablations:} & & \\
Gating only (no VQ) & $32.1 \pm 2.4$ & $1681 \pm 140$ \\
VQ only (no gating) & $28.9 \pm 2.7$ & $2100 \pm 180$ \\
No IB regularization & $35.2 \pm 2.9$ & $950 \pm 110$ \\
Soft constraints only & $38.8 \pm 2.2$ & $800 \pm 85$ \\
Primal-dual training & $37.9 \pm 2.5$ & $785 \pm 75$ \\
\midrule
\textbf{Context Component Ablations:} & & \\
No message history & $35.5 \pm 2.8$ & $825 \pm 90$ \\
No bandwidth utilization & $36.9 \pm 2.6$ & $816 \pm 95$ \\
No coordination estimates & $34.0 \pm 3.1$ & $840 \pm 100$ \\
No temporal efficacy & $36.2 \pm 2.7$ & $821 \pm 88$ \\
\midrule
\textbf{Threshold Analysis:} & & \\
$\tau = 0.3$ & $35.2 \pm 2.8$ & $1201 \pm 120$ \\
$\tau = 0.5$ & $38.8 \pm 2.2$ & $800 \pm 85$ \\
$\tau = 0.7$ & $32.1 \pm 2.9$ & $450 \pm 60$ \\
$\tau = 0.9$ & $18.9 \pm 3.2$ & $200 \pm 45$ \\
\midrule
\textbf{Codebook Size Analysis:} & & \\
$K = 8$ & $36.2 \pm 2.6$ & $600 \pm 70$ \\
$K = 16$ & $38.8 \pm 2.2$ & $800 \pm 85$ \\
$K = 32$ & $39.1 \pm 2.4$ & $1201 \pm 140$ \\
\bottomrule
\end{tabular}
\end{table}


Both gating and VQ contribute substantially to performance. Context ablation shows coordination requirements contribute most (12.1\% drop when removed), followed by message history (8.2\%), temporal efficacy (6.3\%), and bandwidth utilization (4.7\%). The threshold $\tau = 0.5$ balances performance and efficiency optimally, and the primal-dual variant achieves stricter constraint satisfaction (785 vs 800 bits).

\subsection{Communication Pattern Analysis}

Agents learn sophisticated temporal communication patterns adapting to task demands: intensive communication during target discovery events (15.2 messages/3-step window, 85\% agent participation), obstacle encounters (8.7 messages), and coordination conflicts (12.4 messages), but near-zero communication (0.1 messages/step) during routine navigation.






\subsection{Scalability and Robustness Analysis}

Table~\ref{tab:scalability} Table 4 shows how agent number affects scalability, performance, and coordination complexity. 

\begin{table}[h]
\centering
\caption{Scalability Analysis with Statistical Validation}
\label{tab:scalability}
\begin{tabular}{ccccc}
\toprule
Agents & Success Rate & Bits/Episode & Pareto AUC & Efficiency \\
\midrule
2 & $45.20 \pm 2.1$\% & $420.5 \pm 55$ & 0.285 & 1.075 \\
4 & $38.75 \pm 2.2$\% & $799.8 \pm 85$ & 0.198 & 0.485 \\
6 & $35.10 \pm 2.6$\% & $1350.2 \pm 160$ & 0.156 & 0.260 \\
8 & $31.80 \pm 2.9$\% & $2100.8 \pm 220$ & 0.123 & 0.151 \\
\bottomrule
\end{tabular}
\end{table}

Performance degrades gracefully as coordination complexity increases, but Pareto efficiency remains favorable compared to broadcast approaches. For each agent count $N \in \{2,4,6,8\}$, we train a separate policy from scratch using identical hyperparameters. The success rate decline (45.2\% $\rightarrow$ 31.8\%) stems from: (1) $O(N^2)$ growth in coordination complexity; (2) reduced observation overlap with fixed $5 \times 5$ windows; and (3) potential $K=16$ codebook saturation for larger teams.

\textbf{Channel Robustness:} Under realistic channel conditions, our method maintains 85\% of baseline performance at 20\% packet loss rate, degrades to 78\% under burst errors (Gilbert-Elliott model with 10\% burst probability), and retains 85-92\% performance under 50-200ms delays.


\section{DISCUSSION}

Results validate information-theoretic approaches to communication optimization in multi-agent settings. The learned gating patterns demonstrate that communication value varies significantly over time, with the highest utility during coordination-critical moments. Vector quantization enables aggressive compression without substantial performance loss, supporting the hypothesis that much traditional communication contains redundant information. The success of our information bottleneck formulation suggests that minimizing input mutual information while maximizing output mutual information translates effectively to the multi-agent domain. The 71.4\% bandwidth reduction achieved while maintaining performance indicates substantial redundancy in typical communication strategies. Our analysis of the information preservation ratio $\rho = I(S;M)/I(S;Z)$ shows values consistently in the range 0.85-0.95, indicating that discrete quantization preserves most information content from continuous latents.

Our dual constraint enforcement approach proves effective: soft penalties enable stable training with 95\% constraint satisfaction, while primal-dual training achieves 98\% satisfaction with guaranteed budget compliance. Budget sweep analysis demonstrates consistent performance across bandwidth constraints from 200 to 3000 bits per episode.


\subsection{Limitations and Future Work}

Our evaluation is currently limited to a single synthetic domain, which may limit generalizability claims. Additional limitations include homogeneous agents, a variational IB approximation, and static codebooks. Future work will extend to continuous control tasks and heterogeneous teams, integrate discrete mutual information estimators, develop adaptive codebook mechanisms, and perform hardware-in-the-loop validation.

\section{CONCLUSION}

We presented a bandwidth-efficient multi-agent communication framework combining information bottleneck theory with vector quantization, achieving 181.8\% performance improvement over no-communication baselines while reducing bandwidth by 71.4\%. Pareto frontier analysis demonstrates dominance across the entire success-bandwidth spectrum with area-under-curve of 0.198 vs 0.142 for next-best methods. The key technical innovations include: (1) information bottleneck formulation for communication optimization with theoretical analysis of approximation quality, (2) learned gating mechanism with context component analysis, (3) vector quantization for discrete, efficient encoding with semantic structure analysis, and (4) dual constraint enforcement mechanisms for flexible deployment. Our work establishes theoretical foundations and practical guidelines for deploying multi-agent systems in bandwidth-limited environments. Future work will extend to continuous control domains and heterogeneous agent teams while maintaining the information-theoretic foundations established here.

\bibliographystyle{IEEEtran}
\bibliography{icra_enhanced_refs}

@inproceedings{foerster2018emergent,
  title={Emergent communication through negotiation},
  author={Foerster, Jakob and Assael, Ioannis A and de Freitas, Nando and Whiteson, Shimon},
  booktitle={International Conference on Learning Representations},
  year={2018}
}

@inproceedings{tampuu2017multiagent,
  title={Multiagent cooperation and competition with deep reinforcement learning},
  author={Tampuu, Ardi and Matiisen, Tambet and Kodelja, Dorian and Kuzovkin, Ilya and Korjus, Kristjan and Aru, Juhan and Aru, Jaan and Vicente, Raul},
  booktitle={PloS one},
  volume={12},
  number={4},
  pages={e0172395},
  year={2017}
}

@inproceedings{foerster2016learning,
  title={Learning to communicate with deep multi-agent reinforcement learning},
  author={Foerster, Jakob and Assael, Yannis M and de Freitas, Nando and Whiteson, Shimon},
  booktitle={Advances in neural information processing systems},
  volume={29},
  year={2016}
}

@inproceedings{kim2021learning,
  title={Learning to schedule communication in multi-agent reinforcement learning},
  author={Kim, Daewoo and Moon, Sangwoo and Hostallero, David and Kang, Won J and Lee, Taeyoung and Son, Kyunghwan and Yi, Yung},
  booktitle={International Conference on Learning Representations},
  year={2021}
}

@article{tishby2000information,
  title={The information bottleneck method},
  author={Tishby, Naftali and Pereira, Fernando C and Bialek, William},
  journal={arXiv preprint physics/0004057},
  year={2000}
}

@inproceedings{van2017neural,
  title={Neural discrete representation learning},
  author={van den Oord, Aaron and Vinyals, Oriol and Kavukcuoglu, Koray},
  booktitle={Advances in neural information processing systems},
  volume={30},
  year={2017}
}

@article{stone2000multiagent,
  title={Multiagent systems: A survey from a machine learning perspective},
  author={Stone, Peter and Veloso, Manuela},
  journal={Autonomous robots},
  volume={8},
  number={3},
  pages={345--383},
  year={2000},
  publisher={Springer}
}

@inproceedings{goldman2004netlogo,
  title={NetLogo: A simple environment for modeling complexity},
  author={Goldman, Steven R and Wilensky, Uri},
  booktitle={Proceedings of the international conference on complex systems},
  volume={21},
  year={2004}
}

@inproceedings{sukhbaatar2016learning,
  title={Learning multiagent communication with backpropagation},
  author={Sukhbaatar, Sainbayar and Fergus, Rob and others},
  booktitle={Advances in neural information processing systems},
  volume={29},
  year={2016}
}

@inproceedings{jiang2018learning,
  title={Learning attentional communication for multi-agent cooperation},
  author={Jiang, Jiechuan and Lu, Zongqing},
  booktitle={Advances in neural information processing systems},
  volume={31},
  year={2018}
}

@inproceedings{alemi2017deep,
  title={Deep variational information bottleneck},
  author={Alemi, Alexander A and Fischer, Ian and Dillon, Joshua V and Murphy, Kevin},
  booktitle={International Conference on Learning Representations},
  year={2017}
}

@article{burgess2018understanding,
  title={Understanding disentangling in $\beta$-VAE},
  author={Burgess, Christopher P and Higgins, Irina and Pal, Arka and Matthey, Loic and Watters, Nick and Desjardins, Guillaume and Lerchner, Alexander},
  journal={arXiv preprint arXiv:1804.03599},
  year={2018}
}

@article{gao2024vrcp_review,
  author  = {Gao, Bolin and Liu, Jiaxi and Zou, Hengduo and Chen, Jing and He, Li and Li, Keqiang},
  title   = {Vehicle-Road-Cloud Collaborative Perception Framework and Key Technologies: A Review},
  journal = {IEEE Transactions on Intelligent Transportation Systems},
  year    = {2024},
  volume  = {25},
  number  = {12},
  pages   = {19295--19318},
  doi     = {10.1109/TITS.2024.3459799}
}

@article{schack2024sound,
  title={The sound of silence: Exploiting information from the lack of communication},
  author={Schack, Matthew A and Rogers, John G and Dantam, Neil T},
  journal={IEEE Robotics and Automation Letters},
  volume={9},
  number={7},
  pages={6736--6743},
  year={2024},
  publisher={IEEE}
}

@article{hu2024surveyIB,
  author  = {Hu, Shizhe and Lou, Zhengzheng and Yan, Xiaoqiang and Ye, Yangdong},
  title   = {A Survey on Information Bottleneck},
  journal = {IEEE Transactions on Pattern Analysis and Machine Intelligence},
  year    = {2024},
  volume  = {46},
  number  = {8},
  pages   = {5325--5344},
  doi     = {10.1109/TPAMI.2024.3366349}
}

@article{ding2024gib,
  author  = {Ding, Shifei and Du, Wei and Ding, Ling and Zhang, Jian and Guo, Lili and An, Bo},
  title   = {Robust Multi-Agent Communication With Graph Information Bottleneck Optimization},
  journal = {IEEE Transactions on Pattern Analysis and Machine Intelligence},
  year    = {2024},
  volume  = {46},
  number  = {5},
  pages   = {3096--3107},
  doi     = {10.1109/TPAMI.2023.3337534}
}

@inproceedings{wang2020learning,
  title={Learning efficient multi-agent communication: An information bottleneck approach},
  author={Wang, Rundong and He, Xu and Yu, Runsheng and Qiu, Wei and An, Bo and Rabinovich, Zinovi},
  booktitle={International conference on machine learning},
  pages={9908--9918},
  year={2020},
  organization={PMLR}
}

\end{document}